\documentclass[letterpaper, 10 pt, conference]{ieeeconf}  

\IEEEoverridecommandlockouts                              

\usepackage{cite}
\usepackage{graphicx}
\usepackage{amssymb}
\usepackage{amsmath}
\usepackage{subfig}
\usepackage{bbding}
\usepackage{hyperref}

\title{\LARGE \bf
Pedestrian Trajectory Prediction Using Dynamics-based Deep Learning
}

\author{Honghui Wang$^{1}$, Weiming Zhi$^{2}$,  Gustavo Batista$^{3}$, Rohitash Chandra$^{1}$ 
\thanks{Correspondence to: H. Wang,  {\tt\small z5380423@ad.unsw.edu.au} and R. Chandra,  {\tt\small rohitash.chandra@unsw.edu.au}}
\thanks{$^{1}$ Transitional Artificial Intelligence Research Group, School of Mathematics and Statistics, UNSW Sydney, Sydney, Australia}%
\thanks{$^{2}$ Robotics Institute, Carnegie Mellon University, Pittsburgh, USA}
\thanks{$^{3}$ School of Computer Science and Engineering, UNSW Sydney, Sydney, Australia}
}

\begin{document}

\maketitle
\thispagestyle{empty}
\pagestyle{empty}

\begin{abstract}

Pedestrian trajectory prediction plays an important role in autonomous driving systems and robotics. Recent work utilizing prominent deep learning models for pedestrian motion prediction makes limited a priori assumptions about human movements, resulting in a lack of explainability and explicit constraints enforced on predicted trajectories. 
We present a dynamics-based deep learning framework with a novel asymptotically stable dynamical system integrated into a Transformer-based model. We use an asymptotically stable dynamical system to model human goal-targeted motion by enforcing the human walking trajectory, which converges to a predicted goal position, and to provide the Transformer model with prior knowledge and explainability. Our framework features the Transformer model that works with a goal estimator and dynamical system to learn features from pedestrian motion history. The results show that our framework outperforms prominent models using five benchmark human motion datasets.

\end{abstract}

\section{INTRODUCTION}

Human motion analysis is required to develop safe human-human and human-robot interaction systems. Human (pedestrian) trajectory prediction plays an important part in human motion analysis \cite{rudenko2020human} and is widely used in various fields such as autonomous driving systems \cite{leon2021review,huang2022survey} and robot navigation \cite{zhi2021anticipatory}.  
Earlier research on predicting human movements began from physics-based methods, such as the social force model \cite{helbing1995social} and the constant velocity/acceleration model \cite{schubert2008comparison}. In recent years, deep learning models have become appealing for pedestrian (human) trajectory prediction with the public availability of large-scale data \cite{sighencea2021review}. In these approaches, pedestrian motion patterns extracted from trajectories of previously observed humans are used to predict human future movements. 

Deep learning models for this problem have evolved from recurrent neural networks (RNNs) and their variants \cite{schmidhuber2015deep} to Transformer models \cite{vaswani2017attention}. Deep learning models the problem as a spatiotemporal sequence for predicting future positions \cite{zhi2021probabilistic}. In particular, RNNs such as the long short-term memory (LSTM) \cite{hochreiter1997long} model have been widely used to represent the temporal and spatial information of human movements in different scenes/situations  \cite{alahi2016social,xue2018ss,vemula2018social,cheng2021amenet}. The Transformer-based models have shown better prediction accuracy than LSTM-based approaches by capturing temporal dependency directly and spatial information of human movements indirectly \cite{yu2020spatio,giuliari2021transformer,zhou2022ga}.  

Recent efforts have also found notable performance improvements by encoding goals (position or desired location) in the deep neural network with historically observed trajectories \cite{zhao2021you}. Encoding goal information reduces the uncertainty in human motion since humans mostly walk towards a predefined destination. An approach is to model the relationship between semantic goals via computer vision methods, such as right-turn, to achieve trajectory prediction \cite{rasouli2019pie} \cite{rhinehart2019precog}. The other approach is to use the goal position derived by the expert system or sampling in neural network models \cite{dendorfer2020goal,mangalam2020not,zhao2021you}.

Deep learning methods have shown state-of-the-art performance in forecasting human trajectories \cite{tan2022review}. However,
these are considered black-box models that lack explainability, making it unclear why the predicted trajectories have the given shapes \cite{korbmacher2022review}. In addition, limited a priori assumptions about human movements in deep learning models make them lack constraints explicitly enforced when predicting desired trajectories \cite{zhi2021probabilistic}. Explainability and explicit constraints are essential for autonomous entities since these allow autonomous entities to analyze their behaviour, attribute responsibility, and engender trust \cite{pasquale2017toward}. The lack of explainability and the explicit enforcement of known rules of desired predicted trajectories can be addressed by introducing prior knowledge into the deep learning models\cite{zhi2021probabilistic,beckh2023harnessing}.

\begin{figure}[htbp!]
\vspace{8pt}
\centering
  \includegraphics[width=.35\textwidth]{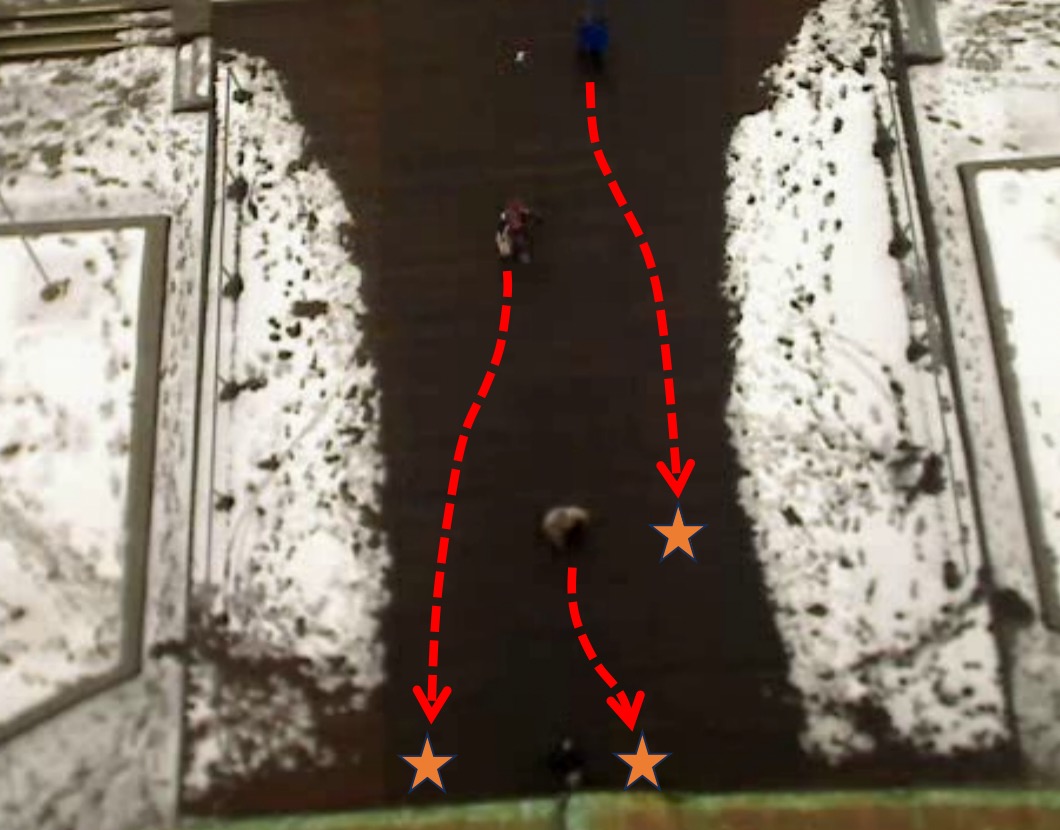}
\caption{Goal-driven pedestrian trajectories where the pedestrian arrives around a predetermined goal.}
\label{fig:1}       
\vspace{-10pt}
\end{figure}

In this paper, we propose a dynamics-based deep learning framework by integrating a novel asymptotically stable dynamical system into a Transformer model. We use the framework to model human motion and can explicitly enforce the human trajectory to converge to the equilibrium point \cite{lyapunov1992general}; i.e. the destination of the pedestrian in the domain of pedestrian trajectory prediction. This property is not only in line with the goal-targeted feature of human movements, as depicted in Fig. \ref{fig:1}, but also can provide the Transformer model in our framework with prior knowledge and explainability. In addition, our framework learns properties of human motion such as continuity, smoothness and boundness \cite{flash1985coordination}, and the temporal and spatial properties of human movements via the Transformer model. 

Specifically, we utilise the goal estimator from Zhao et al. \cite{zhao2021you} to estimate the endpoint in advance for our proposed dynamical system. Then, we use the Transformer-based model -- Spatio-Temporal grAph tRansformer (STAR)\cite{yu2020spatio} to learn an asymptotically stable dynamical system that is behaviorally suitable to human motion to predict the future human walking trajectory more precisely. In summary, our main contribution is to integrate asymptotically stable dynamical systems within a Transformer model to model the human goal-directed motion by enforcing the human walking trajectory to converge to a predicted goal position, thereby providing the Transformer model with prior knowledge and explainability.

\section{Dynamics-based deep learning}

\subsection{Problem formulation}
The main task is to forecast the correct future human motion trajectories during time steps $t_{obs+1}$ to $t_{end}$ given the previously observed trajectories of $M$ pedestrians in 2D environments for $1$ to $t_{obs}$ time steps. In this paper, unless otherwise specified, we denote $i^{th}$ pedestrian trajectory $\xi_{i}(t'', p_{i}(t'))$ starting from $t'$ to $t''$ by a set of $\{p_{i}(t)\}|_{t=t'}^{t=t''}$, where $p_{i}(t) = (x_{i}(t), y_{i}(t)) \in \mathbb{R}^{2}$ is the position of $i^{th}$ pedestrian at time $t \in \{t', t'+1, \ldots , t''\}$, and $i \in \{1,2, \ldots,M\}$.

\subsection{Framework}

\begin{figure}[htbp!]
\vspace{4pt}
\centering
  \includegraphics[width=0.47\textwidth]{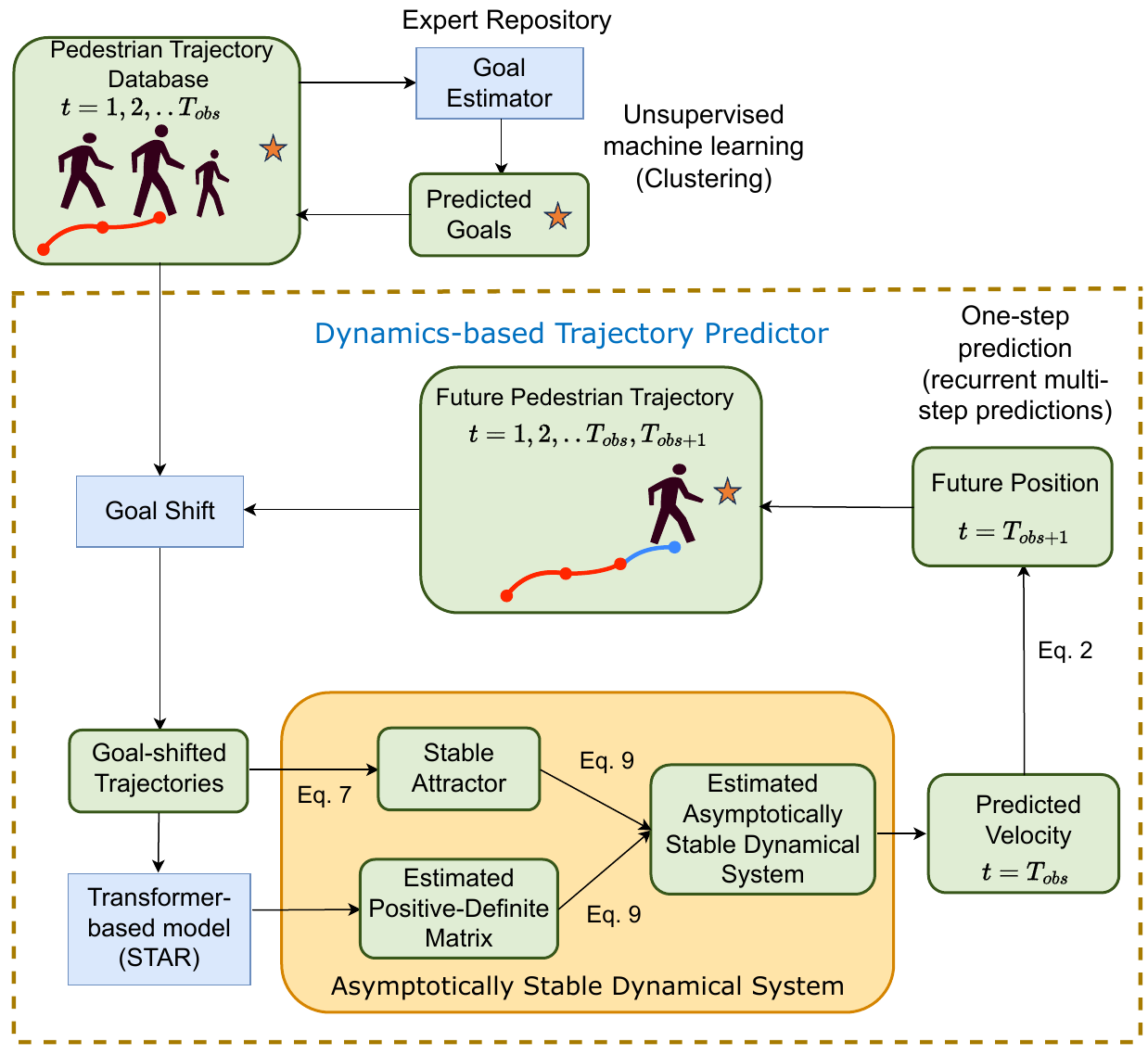}
\caption{We train the Transformer-based model to estimate the positive-definite matrix to predict trajectories. The red line represents the past pedestrian trajectory, while the blue line represents the predicted pedestrian trajectory. The figure shows only a one-step prediction, and the prediction at $T_{obs+1}$ is added back to the past pedestrian trajectory for future steps that are predicted recurrently. The dotted box indicates the dynamics-based trajectory predictor.}
\label{fig:2}       
\vspace{-10pt}
\end{figure}

In our proposed dynamics-based deep learning (DDL) framework (Fig. \ref{fig:2}), we utilise the end position inferred by a goal estimator using unsupervised machine learning (clustering) on an expert repository consisting of the several most similar trajectories to a target trajectory. Our novel asymptotically stable dynamical system and the estimated end position attempt to model human goal-targeted motion representation. We then use the goal shift component to obtain goal-shifted trajectories to estimate the stable attractor. We also use it as an input to the Transformer-based model to estimate the positive-definite matrix. The stable attractor and positive-definite matrix are part of our novel asymptotically stable dynamical system used to predict velocity and one step-ahead pedestrian trajectory. The DDL framework requires a large dataset of pedestrian trajectories that is represented by a set of points in space featuring the location ($x$ and $y$ coordinates). Note that the trajectories will be of varying lengths depending on the pedestrian journey. We use a sliding window approach where we select $w$ input data points from trajectories of varied lengths to make a one-step ahead prediction using DDL. We add the prediction back to the past pedestrian trajectory for the next prediction, which denotes a recurrent approach for multi-step ahead prediction. For instance, we use $w=8$ frames (points) as the input window to predict 12 frames, recurrently. 

\subsection{Human goal-targeted motion representation}

The $i^{th}$ human motion, represented by a mapping between the position $p_{i}(t)\in \mathbb{R}^{2}$ and its velocity $v_{i}(t)\in \mathbb{R}^{2}$, can be modelled as a time-invariant first-order dynamical system \cite{zhi2022diffeomorphic}: 
\begin{align}
v_{i}(t) = f(p_{i}(t))
\end{align}
where $f: \mathbb{R}^{2} \rightarrow \mathbb{R}^{2}$ is a non-linear mapping, representing a dynamical system, and $p_{i}(1)$ is the initial position. The trajectory of human motion can be represented as:
\begin{align}
\xi_{i}(t, p_{i}(1)) = p_{i}(1) + \int_{1}^{t} v_{i}(s) ds
\end{align}
where $\xi_{i}(t, p_{i}(1))$ represents the trajectory of $i^{th}$ human motion starting from the initial time.

A system (human motion) $f$ is globally asymptotically stable in a region $S=\mathbb{R}^{2}$ if trajectories starting in $S$ converge to equilibrium points $p_{i}^{\ast} = p_{i}(t_{end}) \in \mathbb{R}^{2}$, a point where the velocity is zero ($f(p_{i}^{\ast}) = 0$), that is
\begin{align}
\lim_{t \to \infty} \xi_{i}(t, p_{i}(1)) = p_{i}^{\ast}, \forall p_{i}(1) \in S
\end{align}

We model human goal-targeted motion representation by our novel asymptotically stable dynamical system and the equilibrium point (end position) inferred by the goal estimator. 

\subsection{Goal estimator}

The goal estimator is composed of three parts: $\gamma$-Soft-Dynamic Time Warping (DTW) \cite{cuturi2017soft} for matching similar trajectories, construction of an expert repository by selecting the $N$ most similar trajectories, and prediction of endpoints by K-means clustering using the expert repository.

\subsubsection{$\gamma$-soft-DTW algorithm}

Unlike the DTW algorithm \cite{sakoe1978dynamic}, the $\gamma$-soft-DTW algorithm is differentiable and widely used in applications related to time series and defined by:
\begin{equation}
	\begin{split}
	D(X, Y) &= DTW_{\gamma}(X, Y) \\
	&= min_{\gamma}\{<A, \Delta(X, Y)>\}
	\end{split}
\end{equation}
where $\Delta(\cdot)$ is the cost function to measure the Euclidean distance between the temporal sequence $X \in \mathbb{R}^{k \times n}$ and the temporal sequence $Y \in \mathbb{R}^{k \times m}$. $A \in \mathbb{R}^{n \times m}$ is the alignment matrix. $<A, \Delta(X, Y)>$ is the inner product of the cost matrix $\Delta(X, Y)$ with the alignment matrix, reflecting the similarity between two temporal sequences. $min_{\gamma}\{\cdot\}$ is $\gamma$-soft-DTW operator, defined as follows:
\begin{align}
 min_{\gamma}(a_{1}, a_{2}, ... a_{n})=\left\{
\begin{array}{rcl}
min_{i<=n}a_{i}       &      & {\gamma=0}\\
-\gamma log\sum_{i=1}^{n}e^{-a_{i}/\gamma}     &      & {\gamma > 0}
\end{array} \right. 
\end{align}

Typically, pedestrians focus on the neighbours’ velocities when navigating crowded environments \cite{kothari2021human}. Our preliminary analysis reported that the velocity is more informative than absolute coordinates. Therefore, we use pedestrian velocity as the input of $\gamma$-soft-DTW algorithm for a similar search between two temporal sequences, and $\gamma=1$, unlike the previous work by Zhao et al. \cite{zhao2021you}.

\subsubsection{Expert repository}

We firstly use $\gamma$-soft-DTW algorithm to calculate the similarity between a target observed trajectory and all observed trajectories from the train set and then select $N$ most similar trajectories to construct the expert repository. We normalize all trajectories in the expert repository for more accurate prediction, as follows:
\begin{equation}
	\begin{split}
    \tilde{\xi}_{i}(t,\tilde{p}_{i}(1)) = \xi_{i}(t,p_{i}(1)) - p_{i}(1)
	\end{split}
\end{equation}
where $\tilde{p}_{i}(t)$ represents the normalized position of the $i^{th}$ pedestrian from the expert expository at time $t \in \{1, 2, .....t_{end}\}$. In our proposed algorithm, $N=100$.

\subsubsection{Goal estimation by K-means clustering}

We utilize the K-means clustering \cite{hartigan1979algorithm} algorithm to cluster all ground truth endpoints of the normalized trajectories from the expert repository. We consider $K=20$ cluster centers as potential predicted endpoints. Then, we select the cluster center with a minimum distance from the ground truth endpoint on the tested trajectory as the normalized estimated endpoint, which follows the protocol of estimating end positions used in trajectory prediction proposed elsewhere \cite{mangalam2020not}. Finally, we add the initial position of the tested trajectory to the normalized estimated endpoint to restore the original estimated endpoint. 

\subsection{A novel asymptotically stable dynamical system}
Pedestrians typically move towards a predefined destination, and human motion trajectories are generally irregular. To represent human movements more precisely, we propose a novel dynamical system that guarantees stability property based on a stable attractor and natural gradient descent \cite{rana2020euclideanizing}.

The stable attractor in Euclidean space is used in the gradient descent dynamical system to guarantee stability property. The stable attractor is shown as follows:
\begin{align}
\Phi(p_{i}(t)) = ||p_{i}(t) - p_{i}^{\ast}||_{2}
\end{align}
where $p_{i}^{\ast}$ is the equilibrium point, indicating the destination of the $i^{th}$ pedestrian.

The gradient descent dynamical system that uses the stable attractor is shown as follows:
\begin{align}
v_{i}(t) :=  -\nabla_{p_{i}(t)}\Phi(p_{i}(t))
\end{align}
This dynamical system can be guaranteed to be stable since the stable attractor is a valid Lyapunov function \cite{vidyasagar2002nonlinear} for this system. Geometrically, the human motion trajectory generated by this stable dynamical system is straight line with the unit-velocity length and converges to the equilibrium point after a certain time.

Typically, human movement trajectories are generally curved rather than straight lines since they avoid collision with obstacles or pedestrians by altering speed and direction at any time when humans are walking toward their destinations. Hence, it is difficult for a stable gradient descent dynamical system to represent human complex motion. To imitate various types of pedestrian movements, we propose a novel asymptotically stable dynamical system based on the natural gradient descent dynamics \cite{rana2020euclideanizing}, shown as follows:
\begin{align}
v_{i}(t) :=  -P(p_{i}(t))\nabla_{p_{i}(t)}\Phi(p_{i}(t))
\end{align}
Our novel dynamical system can generate the curved trajectory because of natural gradient descent dynamics that are derived from the stable attractor $\Phi(p_{i}(t))$. In addition, it can maintain stability when $P(p_{i}(t))$ is a positive-definite matrix. 

We model human goal-targeted motion representation, but this representation may not be socially acceptable, that is, the lack of social norms like respecting personal space \cite{gupta2018social}. In our stable dynamical system, the walking velocity of a person, which dominates human walking behavior is determined by a positive-definite matrix $P(p_{i}(t))$. To imitate various types of human movement more realistically, $P(p_{i}(t))$ needs to be endowed with certain properties of human motion, such as social norms. 

\subsection{Dynamics-based Trajectory Predictor}
The irregularity of human movements makes it challenging to determine the positive-definite matrix $P(p_{i}(t))$. Data-driven methods have been widely used to represent complex behaviors of human motion. In our framework, the Transformer-based method (STAR) \cite{yu2020spatio} estimates the positive-definite matrix by learning from past human motion trajectories. Specifically, we first apply the goal shift algorithm on the pedestrian trajectory for encoding goal information in the Transformer model. Secondly, we develop our asymptotically stable dynamical system by estimating the parameters of the positive-definite matrix $P(p_{i}(t))$ via the Transformer-based method for generating the velocity that can predict future positions of pedestrians. Finally, we compute the loss function to imitate the human motion trajectory in the real world as accurately as possible.

\subsubsection{Goal shift}

 We use the whole goal-shifted motion trajectory as an input to the Transformer model, which is slightly different from \cite{zhao2021you} and defined as follows:
\begin{equation}
	\begin{split}
\bar{{\xi}}_{i}(t,\bar{p}_{i}(1)) = \xi_{i}(t,p_{i}(1)) - p_{i}(t_{end})
	\end{split}
\end{equation}
where $\bar{p}_{i}(t)$ represents the goal-shifted position of the $i^{th}$ pedestrian at time $t \in \{1,2....,t_{end}\}$. It can be used to generate the stable attractor in our dynamical system. 

During the training phase, $p_{i}(t_{end}) = (x_{i}(t_{end}), y_{i}(t_{end}))$ is the ground truth endpoint, while goal estimator predicts it during the test phase.

\subsubsection{Training the novel asymptotically stable dynamical system}

We decompose the positive-definite matrix $P(p_{i}(t))$ based on lower–upper (LU) decomposition in our proposed dynamical system (Eq. 9) while keeping the positive-definite property, shown as follows:
\begin{align}
P(p_{i}(t)) = L(p_{i}(t))L^{T}(p_{i}(t)) + \sigma I
\end{align}
where $L(p_{i}(t))$ is a lower triangular matrix. $I$ is the identity matrix. We set $\sigma$ to be $10^{-8}$. 

Human motion trajectories can be represented as two-dimension temporal sequences. It is clear that for any $a(p_{i}(t)), b(p_{i}(t)), c(p_{i}(t))\in \mathbb{R}$ in the lower triangular matrix $L(p_{i}(t))\in \mathbb{R}^{2 \times 2}$, $P(p_{i}(t))\in \mathbb{R}^{2 \times 2}$ can keep the positive-definite property.

Therefore, learning the positive-definite matrix can be transformed into learning the three elements in the lower triangular matrix in Eq. 11. The Transformer-based model $f_{T}$ predicts the positive-definite matrix that shapes the motion by learning three elements in the lower triangular matrix given by, 
\begin{equation}
    f_{T}[\bar{{\xi}}_{i}(t,\bar{p}_{i}(1))] = L(p_{i}(t)) = 
\begin{bmatrix} a(p_{i}(t)) & 0 \\ b(p_{i}(t)) & c(p_{i}(t))\end{bmatrix}
\end{equation} 

Pedestrian future positions can then be predicted by integrating over the velocity derived by our well-trained asymptotically stable dynamical system, as in Eq. 2.

\subsubsection{Loss function}

We optimize a mean squared error (MSE) loss between predicted and ground-truth trajectories during training. In particular, we have,
\begin{align}
\mathcal{L}_{\text{MSE}} = \frac{1}{M * (t_{end}-1)}\sum_{i=1}^{M}\sum_{t=2}^{t_{end}}||p_{i}(t) - \hat{p}_{i}(t)||_{2}^{2}\end{align} 
where $\hat{p}_{i}(t)$ represents the predicted position of the $i^{th}$ pedestrian at time $t$.

\section{Empirical Evaluation}
\subsection{Technical Setup}

The ETH/UCY dataset is a collection of benchmark datasets commonly used for evaluating trajectory prediction algorithms. The dataset is divided into two parts: the ETH dataset \cite{pellegrini2009you} and the UCY dataset \cite{lerner2007crowds}. The ETH dataset includes two scenes (ETH and HOTEL), each with a different number of pedestrians, ranging from 28 to 60 individuals. The UCY dataset includes three scenes (UNIV, ZARA1, ZARA2), each with a different number of pedestrians, ranging from 30 to 94 individuals. The dataset developed by processed pedestrian videos from various settings is formatted to include spatiotemporal data representing variable-length trajectory points ($x$ and $y$ coordinates).

We use two prominent metrics that are commonly used for assessing trajectory prediction accuracy, i.e. the average displacement error (ADE) and final displacement error (FDE), which are defined by

\begin{equation}
ADE = \frac{\sum_{i=1}^{M}\sum_{t=t_{obs} + 1}^{t_{end}} \Vert p_{i}(t) - \hat{p}_{i}(t) \Vert_{2}   }{M \times T} 
\end{equation}

\begin{equation}
FDE = \frac{\sum_{i=1}^{M} \Vert p_{i}(t_{end}) - \hat{p}_{i}(t_{end}) \Vert_{2}   }{M} 
\end{equation}

where $T$ is the number of predicted timesteps, and $\hat{p}_{i}(t)$ represents the predicted position of the $i^{th}$ pedestrian at time $t$.

In the ETH/UCY dataset, we use a leave-one-out approach similar to previous works \cite{gupta2018social}, where the model trains on four datasets and tests on the remaining set. 
We use the same data processing strategy by Yu et al.\cite{yu2020spatio} for training and testing with the same hyperparameters, while the training epoch is set to 200.

\begin{table*}[t]
\centering
\vspace{3pt}
\caption{Results of trajectory prediction using ETH/UCY datasets. We use ADE and FDE metrics to compare our proposed method (DDL) with the methods from the literature. The lower error indicates better performance; the best performance is marked in bold.}
\begin{tabular}{lllllll}
\hline
Method                       & \multicolumn{6}{c}{ADE/FDE with Best-of-20 strategy}                                                                 \\ \hline
Dataset                      & ETH                & HOTEL              & UNIV               & ZARA1              & ZARA2              & Average            \\ \hline
Social-STGCNN \cite{mohamed2020social}               & 0.64/1.11          & 0.49/0.85          & 0.44/0.79          & 0.34/0.53          & 0.30/0.48          & 0.44/0.75          \\
PECNet \cite{mangalam2020not}                      & 0.54/0.87          & 0.18/\textbf{0.24}          & 0.35/0.60          & 0.22/0.39          & 0.17/0.30          & 0.29/0.48          \\
Constant Velocity with noise \cite{scholler2020constant} & 0.43/0.80          & 0.19/0.35          & 0.34/0.71          & 0.24/0.48          & 0.21/0.45          & 0.28/0.56          \\
STAR \cite{yu2020spatio}                        & 0.36/0.65          & 0.17/0.36          & 0.31/0.62 & 0.26/0.55          & 0.22/0.46          & 0.26/0.53          \\
Goal-GAN \cite{dendorfer2020goal}                    & 0.59/1.18          & 0.19/0.35          & 0.60/1.19          & 0.43/0.87          & 0.32/0.65          & 0.43/0.85          \\
AVGCN \cite{liu2021avgcn}                      & 0.62/1.06          & 0.31/0.58          & 0.55/1.20          & 0.33/0.70          & 0.27/0.58          & 0.42/0.82          \\
Social-Implicit \cite{mohamed2022social}             & 0.66/1.44          & 0.20/0.36          & 0.31/0.60          & 0.25/0.50          & 0.22/0.43          & 0.33/0.67          \\
GA-STT \cite{zhou2022ga}                      & 0.51/0.89          & 0.22/0.46          & \textbf{0.29}/0.63          & 0.25/0.55          & 0.20/0.44          & 0.29/0.59          \\
SKGACN \cite{lv2023skgacn}                      & 0.55/0.83          & 0.30/0.50          & 0.39/0.75          & 0.30/0.51          & 0.26/0.45          & 0.36/0.61          \\
FlowChain \cite{maeda2023fast}                      & 0.55/0.99          & 0.20/0.35          & \textbf{0.29/0.54}          & 0.22/0.40          & 0.20/0.34          & 0.29/0.52          \\

DDL                       & \textbf{0.26/0.50} & \textbf{0.15}/0.35 & \textbf{0.29}/0.58          & \textbf{0.16/0.29} & \textbf{0.13/0.22} & \textbf{0.20/0.39}  \\ \hline
\end{tabular}
\vspace{-10pt}
\end{table*}

\subsection{Results and Analysis}

We compare our proposed DDL with a wide range of methods, including Social Spatio-Temporal Graph Convolutional Neural Network (Social-STGCNN) \cite{mohamed2020social},
Predicted Endpoint Conditioned Network (PECNet) \cite{mangalam2020not}, Constant velocity with noise \cite{scholler2020constant}, STAR \cite{yu2020spatio}, Goal-Generative Adversarial Networks (Goal-GAN) \cite{dendorfer2020goal}, Attention Visual Field Constraints Graph Convolutional Networks (AVGCN) \cite{liu2021avgcn}, Social-Implicit \cite{mohamed2022social}, Group-Aware Spatial-Temporal Transformer (GA-STT) \cite{zhou2022ga}, Social Knowledge-guided Graph Attention Convolutional network (SKGACN) \cite{lv2023skgacn}, and FlowChain \cite{maeda2023fast}. 

The Best-of-$N$ strategy has been used for evaluating pedestrian prediction models and also used for related problems such as robot swarms \cite{valentini2017best}. Note that our DDL framework uses eight frames (3.2s) as an observed sequence and twelve frames (4.8s) as a predicted sequence on the ETH/UCY dataset. We use the Best-of-$N$ strategy to test our framework and show minimum ADE and FDE from $N=20$ randomly sampled predicted trajectories. DDL features the Gaussian noise in the Transformer-based model (STAR model \cite{yu2020spatio}) and hence this acts as a form of uncertainty projection in the predictions. We implemented the experiments for DDL using \textit{Katana}\footnote{ \url{https://researchdata.edu.au/katana/1733007}}, a high-performance computing service at UNSW Sydney. We used 1 GPU (Tesla V100-SXM2) and 30GB memory, and it took approximately 1 day to train all datasets. Our DDL Python code can be executed both by CPU and GPU.

We can observe from Table I that DDL performs better than all compared methods on most datasets. Compared with the STAR model\cite{yu2020spatio}, DDL achieves better results for all datasets and has an average increase of 23.1\% in ADE and 26.4\% in FDE. This remarkable improvement suggests that the goal shift method and our asymptotically stable dynamical system jointly contribute significantly to forecasting human movements more accurately. We also notice that DDL reports the same result (ADE) on the UNIV dataset with high crowd density as GA-STT, which considers group interactions in the spatial-temporal Transformer network \cite{zhou2022ga}. However, DDL outperforms 31.0\% and 33.9\% in ADE and FDE, respectively, on average. FlowChain \cite{maeda2023fast} also shows the same ADE but lower FDE on the UNIV dataset. However, DDL still outperforms 31.0\% and 25.0\% in ADE and FDE, respectively, on average. PECNet \cite{mangalam2020not} sets the average endpoint between predicted endpoints sampled from latent future endpoints and ground truth endpoints in the loss function to learn parameters for estimating future endpoints. In the case of the HOTEL dataset where the percentage of pedestrians who are almost waiting or ambling is high, better performance on FDE can be achieved by PECNet via learning scene-matching pedestrian movement patterns. Nevertheless, DDL shows an average improvement of 31.0\% on ADE and 18.8\% on FDE.

\vspace{-3pt}

\subsection{Insights}

We present and review selected trajectories by DDL to gain more insights into the various behaviors of human walking. We would like to ascertain if DDL can imitate a wide range of human movements and predict socially acceptable trajectories.

\subsubsection{Convergence to destination}
\vspace{1pt}
Unlike conventional deep learning methods, where it is hard to explain why the predicted trajectories have given shapes \cite{korbmacher2022review}, DDL can make trajectories converge to endpoints because of the stability property in our proposed dynamical system. We can observe that the convergence property can be reflected by DDL on gradually decreasing velocity as approaching the end position on single and group human movement predictions (Fig. \ref{fig:3}a and Fig. \ref{fig:3}b). This indicates that our asymptotically stable dynamical system in DDL provides the Transformer model with prior knowledge and explainability and imposes explicit constraints on predicted trajectories by integrating stability theory into the deep learning model.

{ \vspace{-1pt}
\setlength{\belowcaptionskip}{-5pt}
\begin{figure}[htbp!]
    \centering
    \subfloat[Single person walking]{\includegraphics[width=0.77\linewidth]{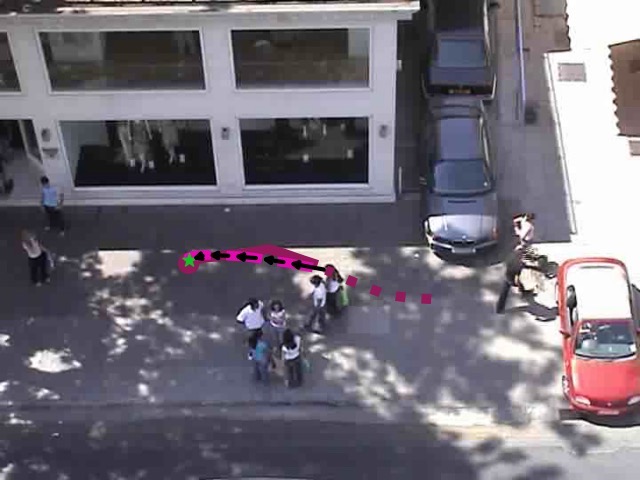}}\\
    \vspace{-7pt}
    \subfloat[Group people walking]{\includegraphics[width=0.77\linewidth]{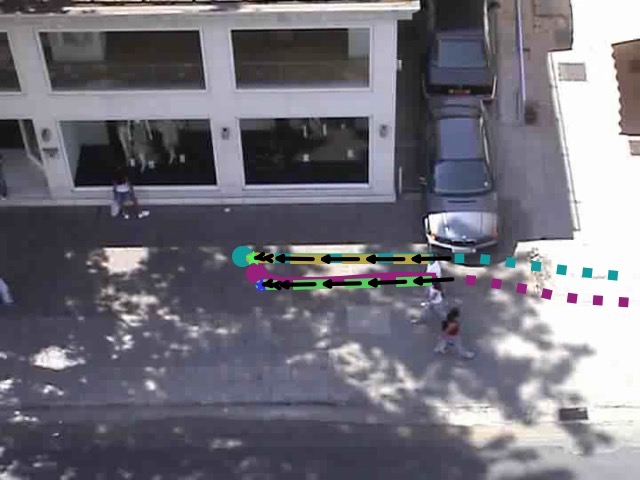}}
    \vspace{-4pt}
    \caption{Trajectory visualization of human goal-targeted walking. The dots represent observed trajectories, discrete lines represent predicted trajectories, and the continuous lines represent ground truth trajectories. The round dots represent ground truth endpoints, and the stars represent predicted endpoints. The arrow represents the velocity, and the length of the arrow indicates the speed.}
     \label{fig:3}
\end{figure}
}

\begin{figure*}[t]
	\centering
	\vspace{5pt}
	\begin{minipage}{0.328\linewidth}
		\centerline{\includegraphics[width=\textwidth]{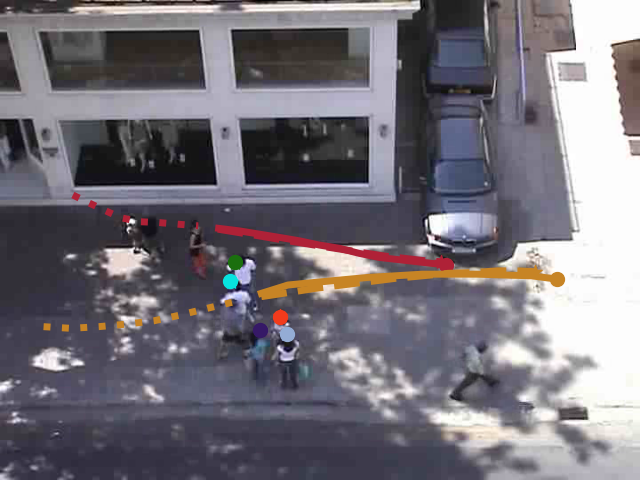}}
		\centerline{(a) Alter velocity}
	\end{minipage}
	\begin{minipage}{0.328\linewidth}
		\centerline{\includegraphics[width=\textwidth]{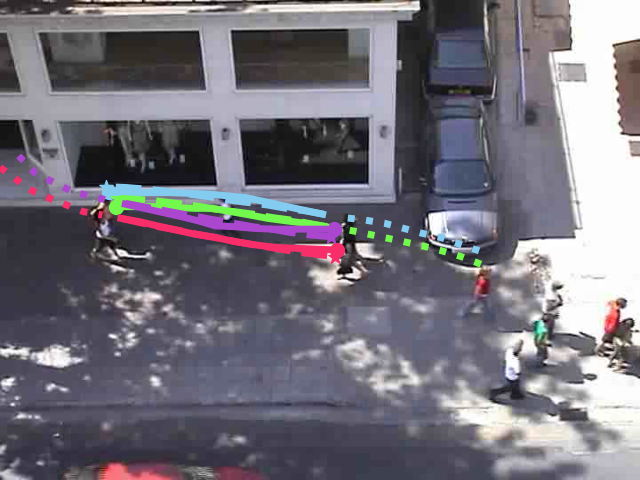}}
		\centerline{(b) Keep social distance}
	\end{minipage}
	\begin{minipage}{0.328\linewidth}
		\centerline{\includegraphics[width=\textwidth]{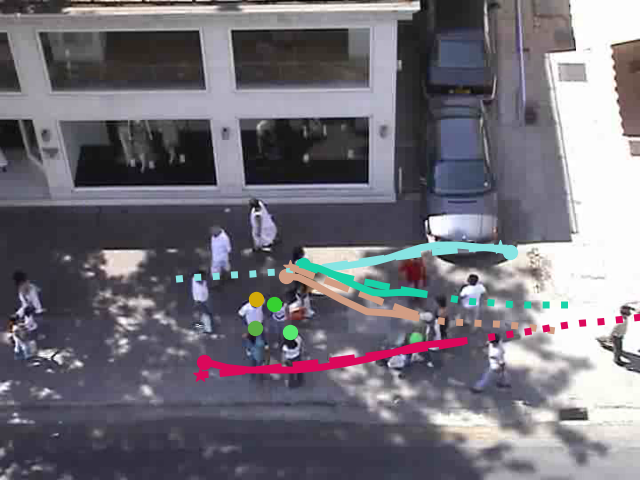}}
		\centerline{(c) Turn around}
	\end{minipage}
	\caption{Trajectory visualization in the case of collision avoidance. The dots represent observed trajectories, discrete lines represent predicted trajectories, and the continuous lines represent ground truth trajectories. The round dots represent ground truth endpoints, and the stars represent predicted endpoints.}
	\label{fig:4}
	\vspace{-3pt}
\end{figure*}

\begin{table*}[t]
\vspace{-3pt}
\centering
\caption{We present results of different combinations of components from DDL using the  ETH/UCY dataset. Component A indicates the goal shift module and Component B indicates our asymptotically stable dynamical system in DDL. Finally, Component C is the STAR model. Note that the lower error indicates better performance; the best is marked in boldface.}
\begin{tabular}{lll|llllll}
\hline
\multicolumn{3}{c|}{Components} & \multicolumn{6}{c}{ADE/FDE with Best-of-20 strategy}           \\ \hline
A         & B        & C        & ETH       & HOTEL     & UNIV      & ZARA1     & ZARA2     & Average   \\ \hline
          &          & \Checkmark         & 0.36/0.65 & 0.17/0.36 & 0.31/0.62 & 0.26/0.55 & 0.22/0.46 & 0.26/0.53 \\
          & \Checkmark         & \Checkmark         & 0.27/0.52 & \textbf{0.15}/\textbf{0.35} & 0.33/0.59 & 0.17/0.31 & 0.14/0.25 & 0.21/0.40 \\
\Checkmark          &         & \Checkmark         & 0.46/0.89 & 0.17/\textbf{0.35} & 0.31/0.64 & 0.17/0.40 & 0.15/0.28 & 0.25/0.51 \\ 
\Checkmark          & \Checkmark         & \Checkmark         & \textbf{0.26}/\textbf{0.50} & \textbf{0.15}/\textbf{0.35} & \textbf{0.29}/\textbf{0.58} & \textbf{0.16}/\textbf{0.29} & \textbf{0.13}/\textbf{0.23} & \textbf{0.20}/\textbf{0.39} \\ \hline
\end{tabular}
\vspace{-10pt}
\end{table*}

\subsubsection{Collision avoidance}

Our results show that DDL can predict human movements that have the behavior of collision avoidance (Fig. \ref{fig:4}). As shown in Fig. \ref{fig:4}a, one person slows down to avoid a collision when they see another person walking forward in front of them. Therefore, our proposed algorithm can imitate the one behavior of collision avoidance by changing the speed.

In Fig. \ref{fig:4}b, when a group of people and another group of people walk in the opposite direction, they keep a certain social distance from each other to avoid collision, indicating that our proposed algorithm can generate the human motion trajectory that can avoid collision by maintaining the social distance.

We can see from Fig. \ref{fig:4}c that when a group of people see another group standing and a person walking in the opposite direction, they alter their respective walking directions to avoid collision with this group. It suggests that DDL can capture the property of changing directions to avoid a collision.

\subsection{Ablation Study}

We conduct extensive ablation studies to verify the influence of each module in DDL. In our ablation study, we separately remove the goal shift module and the novel asymptotically stable dynamical system in DDL. The result of the experiment conducted on ETH/UCY datasets is shown in Table II. Our proposed asymptotically stable dynamical system is the main contribution (19.2\% and 24.5\% increases in ADE and FDE, respectively, on average) by comparing the results shown in the second line with those shown in the first line. In contrast, the goal-shifted processing on the original trajectories also shows a slight improvement in performance in both the STAR model only (both ADE and FDE increase by 3.8\% on average), and the combination of the STAR model and our asymptotically stable dynamical system (4.8\% and 2.5\% increases in ADE and FDE, respectively, on average).

\section{Conclusion and future research}
We presented a dynamics-based deep learning framework for pedestrian trajectory prediction that features a novel asymptotically stable dynamical system to model human goal-targeted movements with the help of a goal estimator. These components provide the Transformer model with prior knowledge and explainability and enforce explicit constraints on predicting trajectories by integrating stability theory into the deep learning model. 
The Transformer-based model provides the dynamical system with insights learned from trajectories that feature human movements, such as collision avoidance in crowded spaces. Our framework performs better than related methods in human walking motion prediction on five benchmark pedestrian walking datasets, i.e. by at least 23.1\% and 18.8\% in ADE and FDE, respectively, on average.

Our framework can predict trajectories with limited motion angle changes only due to the inherent properties of positive definite matrices geometrically. In future research, concepts from diffeomorphism that potentially curve the trajectory at a larger angle can be incorporated to overcome this limitation. In addition, we can use other models apart from the STAR model in our framework and additional datasets to evaluate model capability in different social settings and environments.  

\section*{Code and Data}
 We have released our code and dataset on GitHub. \footnote{\url{https://github.com/sydney-machine-learning/pedestrianpathprediction}}


\bibliographystyle{IEEEtran}
\bibliography{IEEEabrv,cas-refs}

\addtolength{\textheight}{-12cm}   


\end{document}